\documentclass[sn-nature]{sn-jnl}

\usepackage{graphicx}%
\usepackage{multirow}%
\usepackage{amsmath,amssymb,amsfonts}%
\usepackage{amsthm}%
\usepackage[title]{appendix}%
\usepackage{xcolor}%
\usepackage{textcomp}%
\usepackage{manyfoot}%
\usepackage{booktabs}%
\usepackage{algorithm}%
\usepackage{algorithmicx}%
\usepackage{algpseudocode}%
\usepackage{listings}%
\usepackage{comment}
\usepackage{subcaption} 
\usepackage{xspace}
\newcommand{\ie}{i.e.\xspace}

\usepackage{adjustbox}
\usepackage{tabularx} 
\usepackage{float}

\theoremstyle{thmstyleone}%
\newtheorem{theorem}{Theorem}
\newtheorem{proposition}[theorem]{Proposition}%

\theoremstyle{thmstyletwo}%
\newtheorem{example}{Example}%
\newtheorem{remark}{Remark}%

\theoremstyle{thmstylethree}%
\newtheorem{definition}{Definition}%

\begin{document}

\title[Leveraging AI for fine-grained food safety risk forecasting in sparse data conditions]{Leveraging AI for fine-grained food safety risk forecasting in sparse data conditions}


\author[1,2]{\fnm{Dongqi} \sur{Wang}}
\email{wangdongqi@zju.edu.cn}

\author[3]{\fnm{Weiwei} \sur{Chen}}
\email{wchen@business.rutgers.edu}

\author[1,2]{\fnm{Han} \sur{Zhou}}
\email{zhou-hannah@foxmail.com}

\author*[1,2]{\fnm{Weihua} \sur{Zhou}}
\email{larryzhou@zju.edu.cn}

\affil[1]{%
  \orgdiv{School of Management},
  \orgname{Zhejiang University}
}

\affil[2]{%
  \orgdiv{Zhejiang Key Laboratory of Decision Intelligence},
  \orgname{Zhejiang University}
}

\affil[3]{%
  \orgdiv{Department of Supply Chain Management},
  \orgname{Rutgers University}
}


\abstract{Ensuring food safety represents a critical public health challenge, particularly when inspection resources are limited and regional sampling data are sparse. This study proposes a Transformer-based framework capable of forecasting fine-grained, city-level food safety risks by unifying over 11 million inspection records with supplemental demographic, economic, and environmental indicators extracted from the Statistical Yearbook. A three-stage pretraining design leverages partial supervision from the Wilson interval (capturing both safety and risk rankings), together with semi-supervised label refinement, to effectively utilize historical records even when local sample sizes are insufficient. Experimental evaluations on data from 2022 show that the proposed approach outperforms baselines significantly. A subsequent field experiment in collaboration with the Zhejiang Provincial Administration for Market Regulation further demonstrates improved detection rates and more efficient allocation of inspection resources compared to a manually developed plan. Observations of regulatory decision-making reveal a threshold-based heuristic employed by inspectors, hinting that additional training or decision-support interfaces could further enhance the impact of AI-generated risk scores. Overall, these findings underscore that a rigorous integration of large-scale public inspection data, Wilson interval-based confidence modeling, and advanced deep learning can facilitate earlier and more granular identification of food safety threats. By reducing reliance on reactive measures alone, the proposed framework has the potential to advance proactive, data-driven oversight of the global food supply.
}




\maketitle

\section{Main}\label{sec1}
Food safety regulation is a fundamental pillar of public health, safeguarding populations from foodborne illnesses and the associated socioeconomic costs. According to the World Health Organization (WHO), approximately 600 million people, 10\% of the global population, contract foodborne diseases each year, leading to an estimated 420,000 deaths. Children under five are among the most vulnerable, bearing nearly 40\% of the disease burden and accounting for approximately 125,000 deaths annually. In low- and middle-income countries, unsafe food further incurs annual losses of roughly \$110 billion due to reduced productivity and elevated healthcare expenditures~\citep{who-food-safety}.

Despite the critical importance of food safety, regulatory agencies face significant challenges in resource allocation for risk detection and management. For instance, the United States Food and Drug Administration (FDA) is only able to inspect 1–2\% of the 40 million imported food shipments each year, leaving substantial vulnerabilities in oversight \citep{fda-tests}. This shortfall is even more pronounced in nations like China \citep{wordbank}, which features vast food production and consumption volumes but operates with limited inspection resources \citep{jaffee2018safe}. The resulting health risks and economic consequences underscore the urgency of developing more effective strategies for identifying and mitigating food safety hazards worldwide \citep{who-food-safety-2024}.

Early warning systems for food safety risks can play an important role in guiding regulatory authorities to strategically allocate resources, thereby enabling timely detection of hazards and targeted interventions \citep{faoun}. Several such systems are already operational, including the European Commission’s Rapid Alert System for Food and Feed (RASFF), which enables swift information exchange between member countries when food or feed poses a risk to public health \citep{rasff}. Moreover, the WHO has issued an operational guide \citep{ewar}, which offers directions on when and how to implement or strengthen early warning and response mechanisms during food safety emergencies. In the broader public health space, open-source epidemic intelligence has employed the One Health Principle to facilitate early threat detection and subsequent interventions \citep{abdelmalik2018epidemic}. However, most of these existing initiatives remain symptom-based, meaning they only identify risks once threats have already materialized \citep{liu2022automated}. Consequently, a more proactive framework is required—one capable of detecting the signals that precede the emergence of food safety risks. The WHO’s Global Strategy for Food Safety underscores this need, advocating for forward-looking national food safety systems that monitor the drivers and trends leading to foodborne hazards \citep{who-gt}. In line with this strategy, this study focuses on identifying indicators that may signify impending food safety threats and harnessing deep learning models to leverage these signals to support timely interventions.

Specifically, this study aims to develop an early warning system for food safety risks that can guide regulatory agencies in efficiently allocating inspection resources, thereby enabling timely identification and control of hazards. To this end, we employed web crawlers to collect food inspection reports released between 2014 and 2024 by various levels of China’s food safety regulatory bodies, building a dataset of more than 11 million entries \citep{jin2021testing}. These data originated from national and provincial authorities, as well as 209 local agencies under the jurisdiction of 334 municipal-level governments. In addition, based on existing literature and expert opinions, we extracted indicator data from the China Statistical Yearbook \citep{yearbook} that are potentially linked to food safety risks for integration into our model.

\begin{figure}
\begin{center}
\centerline{\includegraphics[scale=0.62]{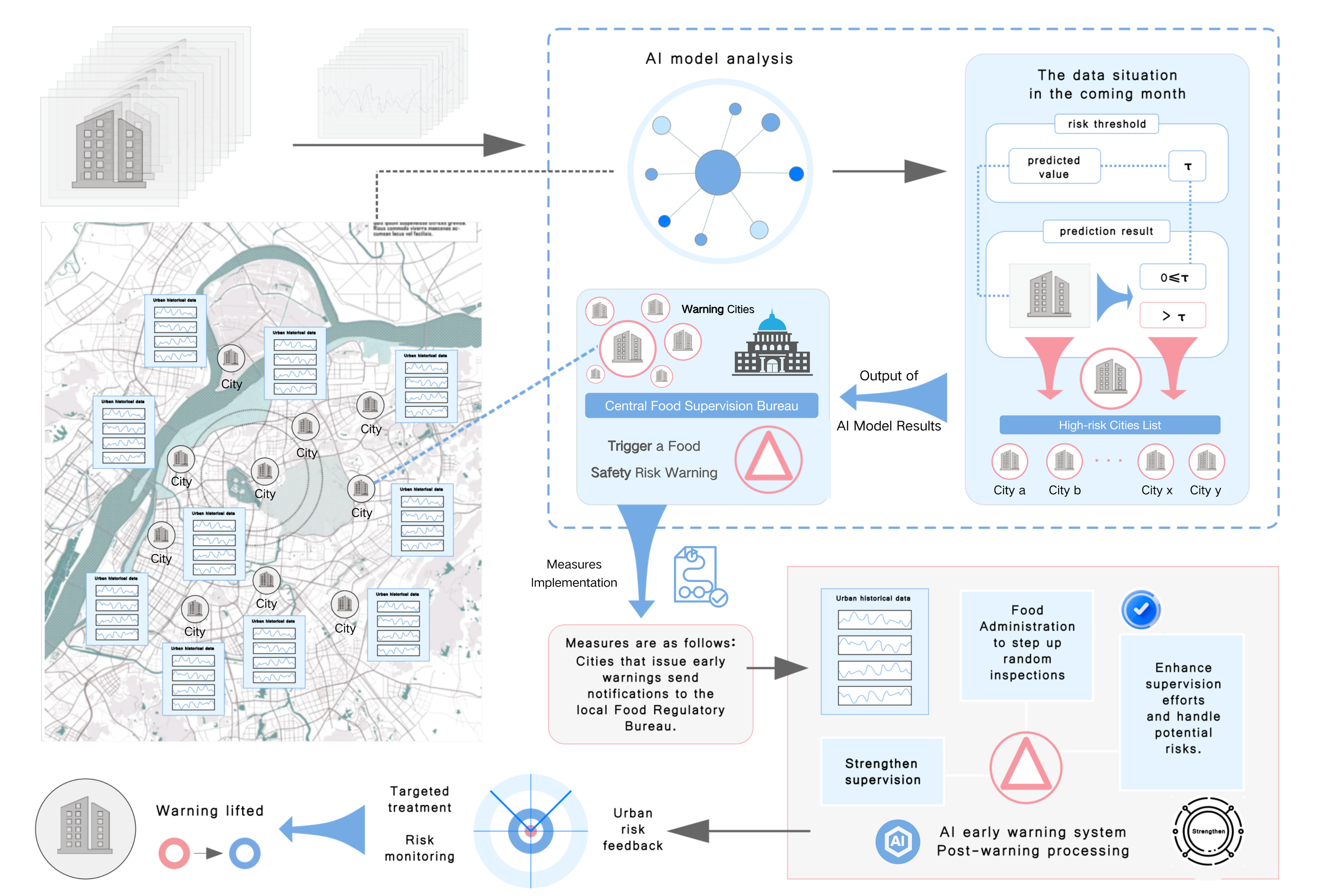}}
\end{center}
\caption{\textbf{Overview of the proposed food early warning system.} When real-time food inspection data from different locations across China are input into our deep learning-based early warning framework, the model predicts the future risk levels of each location and issues warnings for high-risk areas. Due to the centralized nature of China’s food regulation administrative system \citep{wu2018food}, the model’s risk predictions can be directly disseminated to the corresponding local administration for market regulation. From both central and local perspectives, this enables the allocation of additional inspection and supervision resources to high-risk regions in specific risk categories (e.g., food types), achieving intelligent risk reduction.}
\label{fig:system}
\end{figure}

However, as previously noted, the limited availability of sampling resources at a specific geographical location poses a major challenge. While the computed risk assessments at aggregated levels (e.g., the annual failure rate for a single province, defined as the proportion of noncompliant items among all products) can be statistically reliable, they lack the granularity required to guide early warnings and resource allocations. On the other hand, the available data become sparse and statistically insignificant at finer spatiotemporal resolutions (e.g., monthly failure rates for a specific food item in a particular city). To address this dilemma, we leverage the Wilson score to establish a statistically meaningful lower bound for real food safety risks. Unlike conventional point estimates or naive confidence intervals, the Wilson score interval accounts for sampling variability and adjusts for small sample sizes, preventing underestimation of risk when observations are limited or when the observed proportion is near the extremes, thereby enhancing robustness \citep{wilson1927probable}. We then apply a transformer-based model with a three-stage pretraining strategy to effectively integrate multiple risk indicators \citep{vaswani2017attention}. Using fresh food as a case study, the proposed framework generates one-month-ahead early warnings of food safety risks for selected cities in China. This approach yields robust, high-granularity risk forecasting and offers practical value in supporting early risk detection efforts across the country, as shown in Fig.~\ref{fig:system}.

\section{Results}
\subsection{Dataset Construction and Risk Indicators}

This study is based on a self-constructed dataset comprising more than 11 million records of passing and failing results from food adulteration tests conducted by China’s state (central), provincial, and municipal-level Administrations for Market Regulation (AMRs) between 2014 and 2024. These records encompass tests posted by the national AMR, all 31 provincial AMRs, and 273 out of 333 municipal-level AMRs, collectively covering almost all major urban and key agricultural regions in China. Rigorous data cleaning procedures were performed to ensure data reliability, including deduplication to prevent double-counting of tests, assessment of reporting rates per location relative to population size, and checks for missing  data or other anomalies (see \citep{jin2021testing} for details). All reported tests follow uniform technical standards and protocols specified in China’s national testing plan \citep{notice2019}, ensuring methodological consistency across the entire dataset. The AMR agencies are legally mandated to publicly disclose all food test results, and the random sampling policy within each product category further supports the dataset’s representativeness \citep{guidance2011}.

Fresh foods were selected as the primary focus in this study, as they are a major component of the daily diet and are more susceptible to external environmental factors \citep{guidance2011}. After integrating relevant indicators from the China Statistical Yearbook, identified through literature review and expert opinions, data from 191 cities covering a 96-month period (2014–2022) were retained, with some cities excluded due to significant missing data. 

\begin{figure}[H]

\begin{subfigure}[b]{1\linewidth}
    \centering
    \includegraphics[width=\linewidth]{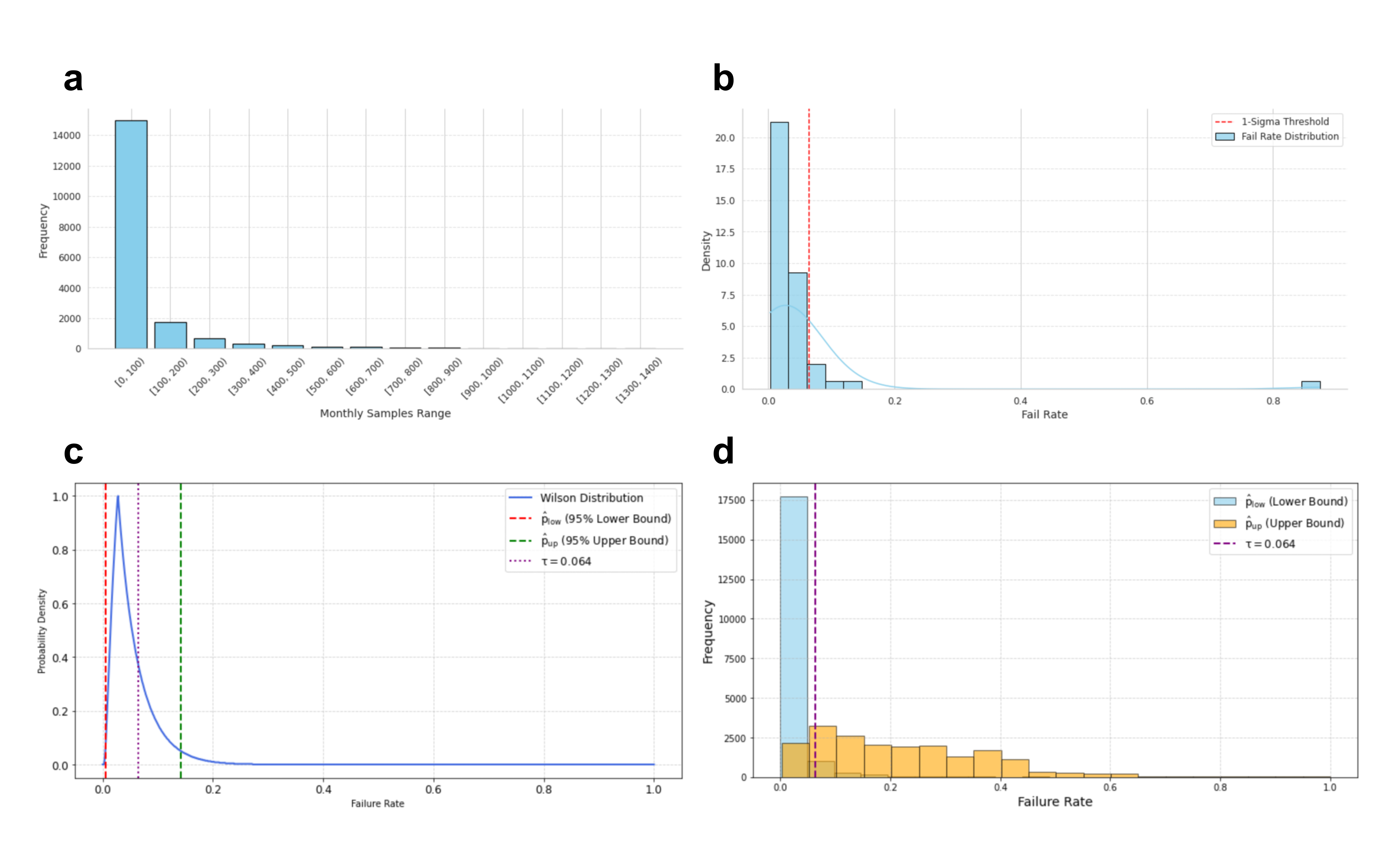}
\end{subfigure}
\hfill



\caption{\textbf{Illustration of sampling data, risk indicator, and Wilson interval analysis.} (a) Description of sampling data, including sample size and distribution parameters. (b) Risk indicator derived from the Wilson score interval, with confidence bounds for defect rate estimation. (c) Probability density function of the Wilson score interval, illustrating uncertainty in failure rate probability. (d) Distribution of Wilson interval bounds (upper and lower) for failure rate across repeated sampling trials.
}
\label{fig:indicator_c}
\end{figure}

While the failure rate is commonly used by regulators to gauge food safety risks, it faces fundamental limitations under small sampling regimes. As illustrated in Fig. \ref{fig:indicator_c}, most city-month pairs in the raw data have fewer than 100 samples. A sample size of 100 results in a margin of error of approximately 3.77\% at a 95\% confidence level. Considering that the overall failure rate of all data is 3.85\% ($p = 0.064$), a margin of error of 3.77\% is quite substantial, rendering the failure rate computed at the city-month level with a small sample size unreliable for meaningful analysis. This is a common challenge across many cities at finer spatiotemporal resolutions for computing food safety risks. 

To mitigate this limitation, we have performed two data processing steps. First, we removed cities with small samples, resulting in a final dataset that aligns with the Cochran sample size formula \citep{cochran1977sampling}; this guideline suggests that a minimum of about 1,400 samples is required to achieve statistically meaningful estimates of failure rates (less than 1\% error at a 95\% confidence level).  Consequently, city-month pairs exceeding 1,400 samples were extracted for subsequent analyses. These records follow a log-normal distribution (Kolmogorov–Smirnov test for log-normality: $p = 0.48$) in Fig. \ref{fig:indicator_c}(a), and we define city-month outliers (one standard deviation above the mean in the log-normal space) as high-risk cases.

Further, we adopt the Wilson score \citep{wallis2013binomial} interval, which provides statistically robust confidence bounds around the observed failure rate. Unlike simple proportion calculations for the failure rate, the Wilson score interval inherently adjusts for small sample sizes by incorporating both the observed data and the uncertainty associated with limited observations, thus offering a more accurate and reliable estimation of the true failure risk. The Wilson score interval has been effectively applied in various fields, such as content ranking algorithms used by popular platforms like Reddit.
The Wilson score interval is defined as follows:
\begin{equation}
  \label{eq:wilson}
  \bigl[\hat{p}_\mathrm{low},\;\hat{p}_\mathrm{up}\bigr]
  \;=\;
  \frac{\hat{p} + \tfrac{z_{\alpha}^2}{2n}
  \;\pm\;
  z_{\alpha}\sqrt{\frac{\hat{p}(1-\hat{p})}{n}
                  + \tfrac{z_{\alpha}^2}{4n^2}}}
       {1 + \tfrac{z_{\alpha}^2}{n}},
\end{equation}
where \(\hat{p} = X/n\) is the observed failure rate, \(n\) is the sample size, and \(z_{\alpha}\) is the critical value for a desired confidence level (e.g., \(z_{0.025}\!=\!1.96\) for 95\% confidence), with $\alpha$ set as 0.025. Using these intervals, we classify city-month samples into three categories: \textbf{low risk} (upper bound below a chosen threshold), \textbf{uncertain} (upper bound above the threshold, yet lower bound remains below it), and \textbf{high risk} (lower bound above the threshold).  Fig. \ref{fig:indicator_c}, provides an illustration of this approach.

While the Wilson interval enhances the reliability of risk assessments under limited sampling, it essentially remains a reactive metric: it interprets past or current testing outcomes. To achieve a proactive risk-warning system, we further develop a deep learning framework that forecasts future food safety risks by integrating historical testing records with various demographic, environmental, and socioeconomic indicators. The subsequent sections detail how the proposed Transformer-based model leverages Wilson interval labels in both ranking and semi-supervised learning paradigms, ultimately facilitating city-level risk prediction before hazards escalate.

\subsection{Overview of transformer model}
Our proposed transformer model aims to predict city-level food safety risks one month in advance, based on multi-dimensional time series data covering historical observations from all available months for each city. Note that while city-level prediction is used to illustrate the proposed model, the granularity can be finer at any region level (e.g., counties or local markets) when relevant data is available in each specific region. 

Formally, the input consists of a $D$-dimensional feature sequence for each city, spanning time steps $1$ to $T-1$. This input passes through $L$ transformer blocks with masked multi-head self-attention to capture both short- and long-term dependencies in the temporal data. The model then produces a latent representation for time step $T$, which is subsequently fed into a classification layer that outputs a binary risk label (low vs. high risk). However, a central challenge arises from the relatively high dimensionality of the input features alongside a limited number of definitive samples: the scarcity of definitively labeled samples—cities that are unambiguously classified as either high or low risk—poses a dilemma. To address this limitation, we enhance the training process via (1) a three-stage pretraining scheme designed to learn robust city representations, and (2) a semi-supervised approach that leverages Wilson score intervals to incorporate “uncertain” samples with soft labels.

\begin{figure}
\begin{center}
\centerline{\includegraphics[scale=0.07]{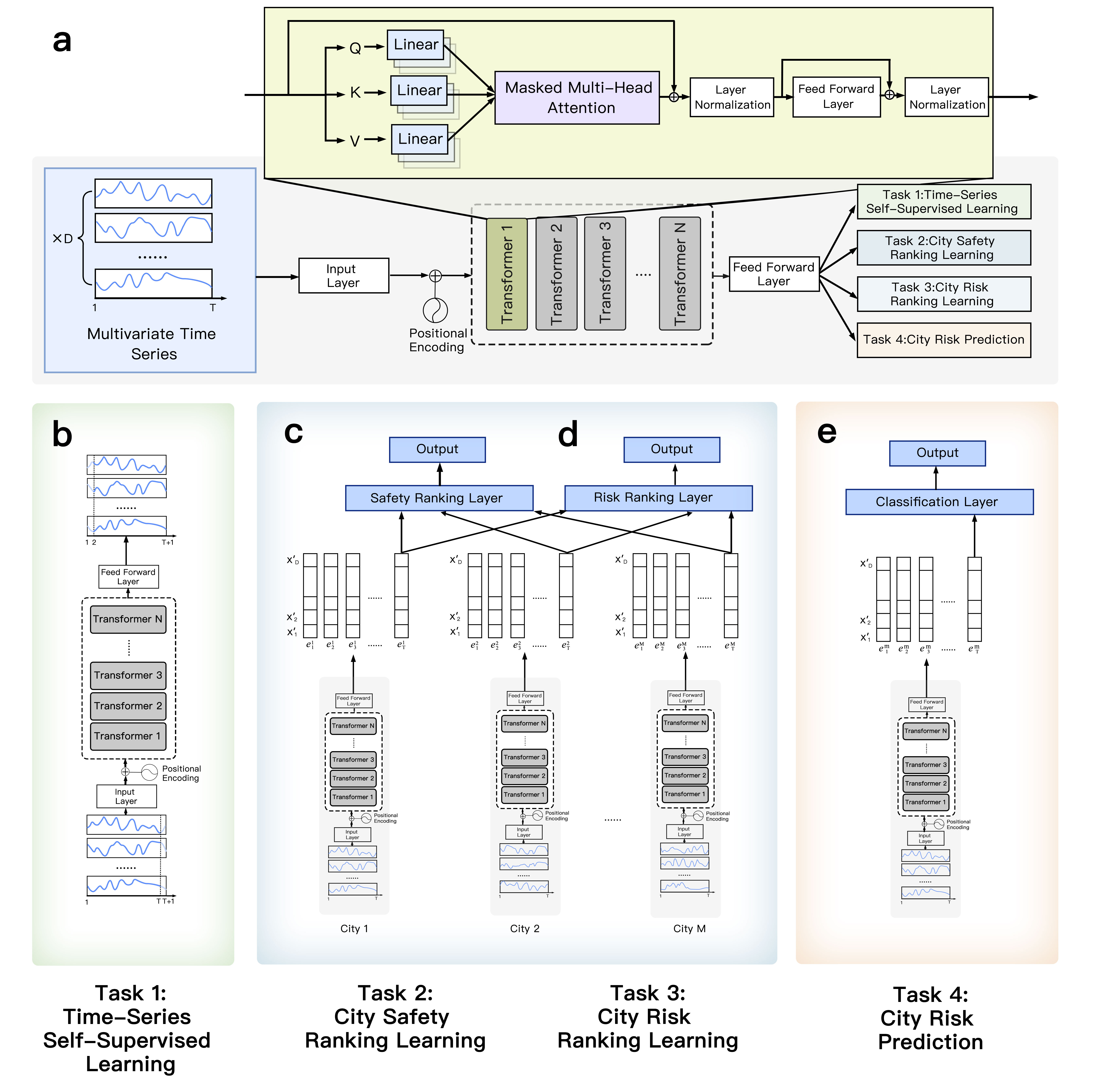}}
\end{center}
\caption{\textbf{Overview of the prediction model architecture.} The model is designed to leverage deep learning technologies for real-time analysis of food inspection data across various regions nationwide, aiming to provide early warnings on potential future food safety risks.(a) The time series-based food safety indicators, serving as input features for AI modeling, are first processed through Transformer blocks with masked multi-head self-attention mechanisms, enabling the model to capture both short- and long-term dependencies while extracting high-level representations. A unique three-stage pre-training framework, integrated with semi-supervised learning, is employed to enhance the model’s performance. (b) In the time-series self-supervised learning stage, it predicts future features from historical data to learn time-series patterns. (c) In the city safety and risk ranking learning stages, the Wilson score is used to rank cities' safety and risk levels, facilitating the model's learning of risk differences among cities. (d) In the semi-supervised learning stage, \textbf{uncertain} samples determined by the Wilson interval are assigned soft labels to enrich training information, and the classification layer ultimately outputs the risk label for the next month, enabling accurate city-level food safety risk prediction.}
\label{fig:transfomer}
\end{figure}

\textbf{Time-Series Self-Supervised Learning}. In the initial pretraining stage, the model aims to reconstruct the features at time $T$ given the $D$-dimensional observations from months $1$ to $T-1$. By letting $T$ vary (e.g., $1, 2, 3, \dots$), each city’s time series can generate multiple training samples without requiring explicit risk labels. This self-supervised objective enables the transformer layers to learn generic temporal patterns and dependencies.

\textbf{City Safety Ranking Learning}. Given the advantages of the Wilson score interval discussed earlier, we apply it specifically to learning city safety rankings. In our setting, we leverage this property to learn a safety ranking across multiple cities at time $T$. Treating the cities’ non-failing (i.e., passing) samples as “positive,” we use the Wilson score to generate a relative ordering of cities by their safety levels. The transformer model thus learns to produce city embeddings that are consistent with these rankings, encouraging it to capture nuances that distinguish safer cities from relatively riskier ones.

\textbf{City Risk Ranking Learning}. Analogous to safety ranking, we also apply Wilson score to rank cities by their failure rates. Here, failing samples are “positive,” and the Wilson score quantifies the lower bound of the failure rate. A city with both a high failure rate and sufficient sample size will yield a relatively higher lower bound. By learning a risk ranking, the model refines its ability to differentiate among cities based on risk severity. Importantly, City Safety Ranking and City Risk Ranking are complementary, as both rely on the confidence intervals derived from the Wilson score, i.e., Wilson interval. Merging these perspectives contributes to a richer representation of city-level food safety profiles.

\textbf{Semi-Supervised Learning via Wilson Score Intervals}. After completing the three-stage pretraining, the model proceeds to the primary classification task: predicting whether a city will be high-risk in the upcoming month. Although some city-month pairs can be definitively labeled as high or low risk (e.g., when the lower bound of the Wilson interval exceeds or remains below a specified threshold), a substantial fraction of samples fall into an “uncertain” zone. To maximize available training signals, we assign soft labels to these uncertain samples by calculating the proportion of the Wilson score distribution that lies above or below the threshold. This approach yields probabilistic labels representing the likelihood of high or low risk. Incorporating these soft labels enriches the training data and allows the model to benefit from the partial information contained in uncertain samples. Ultimately, the resulting classifier gains a more nuanced understanding of risk dynamics, improving its performance in real-world settings where unambiguous data are limited.

\subsection{Results of Comparative Benchmarking Experiments}
\label{sec:Numerical experiment}

We conducted a quantitative evaluation using city-level food safety data from 2022 to assess the effectiveness of our approach. Five models were compared: XGBoost, a tree-based model (similar in type to the one used in \citep{martini2022machine} for food security tasks); LSTM, using a long short-term memory architecture; GRU, relying on gated recurrent units; Transformer, a popular structure in time-series prediction tasks; and the proposed model, referred to as \textbf{Ours}, which employs a Transformer-based structure incorporating Wilson interval-guided pretraining and semi-supervised label assignment.

\begin{figure}[H]

\begin{subfigure}[b]{1\linewidth}
    \centering
    \includegraphics[width=\linewidth]{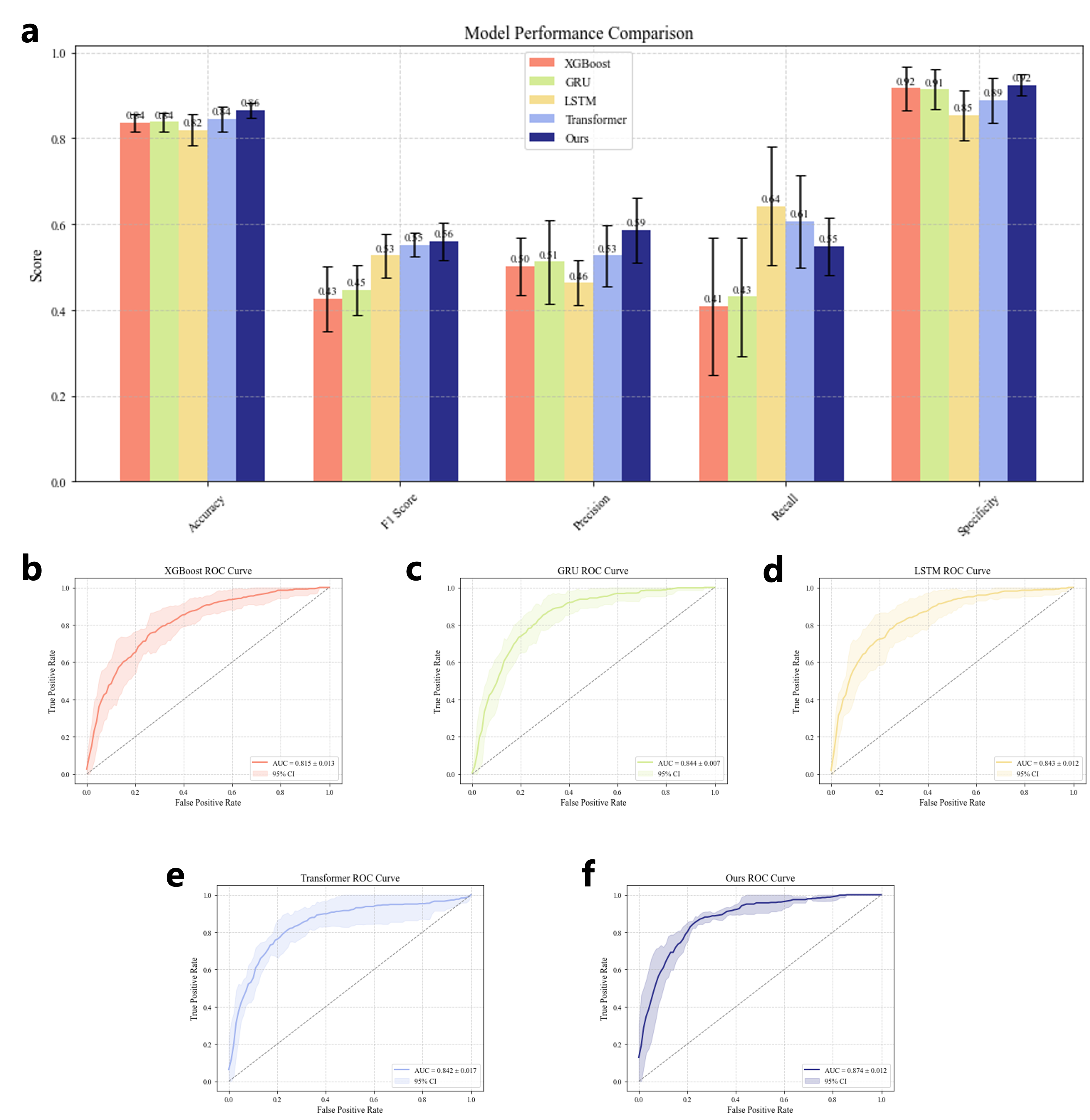}
\end{subfigure}
\hfill
\caption{Overview and detailed comparisons.}
\label{fig:combined}
\end{figure}





Figure~\ref{fig:combined} illustrates the performance of these methods across multiple metrics, and the corresponding receiver operating characteristic (ROC) curves. XGBoost achieves an accuracy of 0.8291, but its $F_{1}$ score and recall are relatively low (0.3692 and 0.3174, respectively), indicating frequent misclassification of high-risk city-months. 

Benefiting from their ability to capture temporal dependencies, LSTM, GRU, and Transformer all outperform the tree-based baselines. Specifically, GRU achieves an accuracy of 0.8308 and an AUC of 0.8048; LSTM achieves 0.8551 and 0.8349, respectively; and Transformer attains 0.8548 in accuracy and 0.8551 in AUC.

The proposed model outperforms all alternatives, reaching an accuracy of 0.8976, an $F_1$ score of 0.6242, and an AUC of 0.9123. Notably, it also attains the highest precision (0.7395), reflecting an enhanced ability to identify genuinely high-risk instances without inflating false positives. Compared with the results of the basic Transformer, the advanced performance gains of the proposed model highlight the benefits of incorporating rich contextual input along with Wilson interval-based pretraining.

\subsection{Results of Experimental Field Experiments}
\label{sec:Empirical experiment}
To further validate the practical value of the proposed food safety risk warning system in real-world scenarios, we collaborated with the Zhejiang Provincial Administration for Market Regulation to conduct a field experiment in Hangzhou during October 2024. The experiment was divided into two groups (each consisting of 100 sampling batches), with the aim of comparing human-directed and AI-guided inspection strategies. In the first group (control group), resource allocation decisions were made manually: decision makers decided how to distribute inspection resources across different categories (e.g., wholesale markets, restaurants, grocery stores) without any algorithmic assistance. In contrast, the treatment group relied on AI risk predictions to guide allocation decisions. Specifically, the AI model, which was trained in earlier numerical experiments, was fine-tuned using monthly food sampling inspection data from various types of sampling sites across cities in Zhejiang Province between January 2022 and September 2024. Based on this updated model, risk predictions were generated for October for different locations in Hangzhou, allowing decision-makers to prioritize potentially high-risk sites more effectively. After finalizing the resource allocation plans, the same inspection team conducted sampling according to the assigned proportions in both groups, enabling a direct comparison of detection rates (i.e., the proportion of noncompliant products).

\begin{table}[!bthp]
    \centering
    \caption{Field Experiment Settings and Results}
    \begin{tabularx}{\textwidth}{l|cc|ccc}
        \toprule
        \multirow{3}{*}{\textbf{Sample Location}} &
        \multicolumn{2}{c|}{Manually-Developed Sample Plan} & 
        \multicolumn{3}{c}{AI-Guided Sample Plan} \\
        \cmidrule(lr){2-3} \cmidrule(lr){4-6}
        & Planned & Failed & Planned & Failed & Prediction \\
        & Batches & Numbers & Batches & Numbers & Results  \\
        \midrule
        Wholesale Market      & 14 & 0 & 14 & 0 & 0.18 \\
        Large Restaurants     &  4 & 0 & 14 & 1 & 0.61 \\
        Others                &  2 & 0 & 15 & 3 & 0.89 \\
        Medium Restaurants    &  8 & 1 & 14 & 2 & 0.73 \\
        Farmers' Market       & 37 & 7 & 15 & 3 & 0.99 \\
        Small Restaurants     &  8 & 1 & 13 & 1 & 0.86 \\
        Small Grocery Store   &  4 & 0 & 15 & 1 & 0.92 \\
        Fresh Food Store      & 23 & 0 & 0 &  0 & 0.04 \\
        Small Supermarket     &  0 & 0 & 0 &  0 & 0.01 \\
        Medium Supermarket    &  0 & 0 & 0 &  0 & 0.01 \\
        Large Supermarket     &  0 & 0 & 0 &  0 & 0.02 \\
        \midrule
        Overall Detection Rate & \multicolumn{2}{c|}{\textbf{9\%}} & \multicolumn{3}{c}{\textbf{11\%}} \\
        \bottomrule
    \end{tabularx}
    \label{tab:FieldExperiment}
\end{table}

\textbf{Risk Detection Efficiency.}
As shown in Table~\ref{tab:FieldExperiment}, the AI-guided inspection plan achieved a detection rate of 11\%, compared to 9\% under the manually developed plan, which is a 2-percentage-point improvement. While this difference may appear modest, a closer comparison between the two plans reveals the practical advantages of AI-guided decision-making. Unlike the manual plan, which relied solely on expert judgment, the AI-guided plan was developed based on continuous-valued prediction scores (logits) indicating the estimated food safety risk level of each location type for October. These scores were provided to regulatory officials as decision-support signals, allowing them to adjust allocations accordingly.

This collaborative approach enabled more targeted and efficient resource use. For example, Fresh Food Stores were predicted to have very low risk (logit = 0.07), and thus received zero inspections under the AI-guided plan. In contrast, the manual plan allocated 23 inspection batches to this category—none of which uncovered any noncompliant samples. This suggests that the AI-guided strategy effectively avoided low-value inspections, conserving limited resources for higher-risk locations. Such differences in allocation patterns demonstrate how AI can complement human expertise by highlighting overlooked inefficiencies in traditional planning, a view echoed in our follow-up interviews with regulatory staff.

Interestingly, the analysis of human decisions suggests that officials may have applied a threshold-based heuristic when interpreting the AI predictions. Specifically, most inspection resources were assigned only to categories with predicted risk logits above 0.1, while locations below this implicit threshold—such as Fresh Food Stores and Large Supermarkets—were largely ignored. Among those exceeding the threshold, however, the allocation appeared nearly uniform, regardless of further differences in predicted risk. This step-function-like allocation behavior may reflect cognitive simplification strategies commonly observed in bounded rationality decision-making, where complex numerical signals are discretized into binary or categorical rules for operational convenience \citep{gigerenzer2011heuristic}.

This behavioral pattern led to a suboptimal use of AI predictions. For example, the AI model predicted a relatively low risk for Wholesale Markets (logit = 0.18), yet 14 inspection batches were still assigned, yielding zero noncompliant findings. In contrast, Farmers’ Markets had the highest predicted risk (logit = 0.89), yet received only 15 batches—identical to the allocation for several lower-risk categories such as Small Grocery Stores (logit = 0.54) and “Others” (logit = 0.69). A more fine-grained interpretation of the prediction scores—i.e., proportionally adjusting inspection intensity based on the relative magnitude of logits—could have significantly enhanced the efficiency of resource use and further improved the detection rate.

Overall, the field experiment highlights the practical value and future potential of our proposed AI-guided risk warning system for food safety inspections. While the current deployment has already led to improved detection outcomes, a closer analysis reveals that regulatory decision-makers tended to adopt a simplified, threshold-like strategy in interpreting AI predictions—prioritizing only those locations with risk logits exceeding a certain value, and allocating resources almost uniformly among them. This behavior is consistent with theories of bounded rationality and heuristic-based decision-making \citep{gigerenzer2011heuristic}, which suggest that humans often convert complex, continuous information into categorical rules to reduce cognitive load. These findings suggest that the proposed AI-guided risk warning system possesses strong potential to improve food safety inspections when paired with appropriately designed human-AI collaboration mechanisms \citep{rahwan2019machine}.

\section{Discussion}
This study addresses the critical challenge of limited inspection resources in food safety regulation by constructing a large-scale dataset and developing a deep learning model that effectively leverages both empirical inspection records and contextual indicators. Integrating over 11 million inspection outcomes with extensive socioeconomic, environmental, and demographic data, we alleviate the difficulties posed by sparse sampling in many regions, where small numbers of tests yield statistically unreliable failure rates. Our proposed Transformer-based model includes three key pretraining stages—time-series self-supervised learning, city safety ranking, and city risk ranking—that exploit Wilson interval guidance. This design substantially increases the amount of usable supervisory information, as demonstrated by the numerical experiments on 2022 data, where the model outperformed multiple baselines significantly.

Beyond computational experiments, the field study conducted in partnership with the Zhejiang Provincial Administration for Market Regulation offers empirical evidence of the model's practical value. Compared to a manually developed inspection plan, the AI-informed approach achieved higher detection rates while allocating fewer resources to low-risk categories. However, closer examination of human decision-making behavior reveals that regulatory officials tended to adopt a simplified, threshold-based interpretation of continuous model outputs. Specifically, they concentrated resources on locations exceeding a certain risk score while distributing these resources in a nearly uniform manner, irrespective of finer-grained variations in predicted risk levels. This pattern of decision-making is broadly consistent with theories of bounded rationality and heuristic-based strategies, in which continuous variables are discretized to reduce cognitive complexity \citep{gigerenzer2011heuristic}.

Although such threshold-based heuristics still led to tangible improvements, they inevitably left some potential gains unrealized. For instance, locations with near-maximal risk scores were allocated roughly the same number of inspections as locations only marginally exceeding the threshold. Additional training or decision-support mechanisms could help regulatory staff move beyond binary interpretations of continuous AI predictions, thereby making more targeted, risk-proportionate allocations. Research on explainable AI and interactive visualization might further enhance trust and comprehension in practice, leading to more adaptive strategies for inspection planning \citep{rahwan2019machine}.

In conclusion, our findings highlight the promise of AI-driven early-warning frameworks for food safety supervision. By unifying extensive public inspection data with contextual features, the proposed method alleviates sampling constraints that frequently hinder accurate assessments of noncompliance rates. Numerical evaluations confirm the model's superior predictive power, while the field deployment in Zhejiang Province underscores its capacity to inform real-world regulatory decisions. Future work may explore additional refinements—such as dynamic thresholding, personalized training for decision-makers, and integration with supplementary risk factors—to further enhance both the model's accuracy and the effectiveness of human-AI collaboration in food safety governance.
\section{Methods}
\subsection{Dataset}
\label{sec:dataset}
In addition to the food safety test data collected from China’s national, provincial, and municipal-level AMRs, we compiled a comprehensive set of contextual and demographic indicators from the China Statistical Yearbook. These auxiliary variables serve to capture potential drivers of food safety risks, such as climate variation, environmental pollution, economic development, and population structure. Specifically, we identified key factors that are theoretically or empirically linked to the incidence of foodborne hazards based on an extensive literature survey.

\textbf{Environmental Indicators}. 
We incorporated monthly temperature records (average, minimum, maximum) spanning 1981–2023, as extreme temperatures can promote microbial proliferation and thus increase food safety risks \citep{checkley2000effects}. We also collected annual data on extreme climate events, which have been associated with emerging foodborne threats \citep{duchenne2021climate}.

\textbf{Economic and Price Indicators}. 
Economic activities were measured by GDP (nominal, real, and GDP deflator) and Consumer Price Index (CPI) from year 2000 onward, reflecting market dynamics that can influence both food production and consumption patterns \citep{iftekhar2020application}. Notably, fluctuations in food prices can alter consumer demand and regulatory focus, thereby affecting the frequency and reporting of food safety incidents \citep{lin2024price}.

\textbf{Environmental Pollution Indices}.
To assess pollution’s impact on agricultural outputs and water/soil quality, key components affecting food safety, we included measures of wastewater, sulfur dioxide (SO\textsubscript{2}), smoke, and dust emissions, as well as pollution removal or treatment rates from year 1990 onward \citep{zhang2015impact, miraglia2009climate}. These indicators capture the complex relationship between industrial activities, environmental degradation, and the safety of food products.

\textbf{City-Level Demographic Data}.
Variables reflecting population size, population growth rate, and urbanization were incorporated to represent how demographic shifts affect both food demand and regulatory capacity \citep{qin2024risk}. Education levels, as evidenced by school enrollment and teacher counts, shed light on public awareness and potential adoption of safe food practices \citep{medeiros2001food}. To gauge agricultural production capacity, we included per-capita arable land area and production levels of major commodities (e.g., vegetables, fruits, meat, and dairy) \citep{carvalho2006agriculture}. Transportation development indicators (i.e., annual freight volume by rail, road, water) were also captured, given that a more advanced logistics infrastructure can both mitigate and accelerate the spread of foodborne risks \citep{han2021comprehensive}.

\textbf{Import-Export Volumes}.
Recognizing that international trade can rapidly disseminate contaminants or pathogens across borders, we collected annual agricultural import and export statistics from year 2001 onward. Such data are especially relevant for evaluating cross-regional and cross-border food safety concerns \citep{ercsey2012complexity}.

\textbf{Geospatial Information}.
Latitude and longitude for each city were recorded to allow spatial analysis of risk distribution. Previous studies have shown that food safety issues often cluster geographically, particularly in highly industrialized or densely populated regions \citep{lee2019does}.

After harmonizing the temporal granularity (monthly or yearly) and resolving discrepancies in reporting formats, we merged these indicators with the AMR test records at the city-month level. We applied rigorous preprocessing to handle missing entries, drawing on multiple imputation and interpolation techniques where possible. Cities that had extensive data gaps were excluded to maintain reliability. Thus, the final data set comprises a rich set of characteristics, environmental, socioeconomic and demographic, aligned by city and month (where applicable) over the 2014–2022 period. This integrated data set supports our subsequent modeling efforts, facilitating a more robust understanding of the contextual factors that influence food safety risk.

\subsection{Model Architecture}
\label{sec:model}

We adopt a Transformer-based encoder to capture temporal dependencies from city-level time series data, ultimately predicting whether a city will be at \emph{high} or \emph{low} food safety risk in the upcoming month. Let \(M\) be the total number of cities. For each city \(m \in \{1,\dots,M\}\), we collect a sequence of \(T\) monthly observations, each of dimension \(D\). Denote this sequence by
\begin{equation}
  \mathbf{X}^m \;=\; 
  \bigl\{\mathbf{x}_1^m,\, \mathbf{x}_2^m,\dots,\mathbf{x}_T^m\bigr\},
  \label{eq:x}
\end{equation}
where \(\mathbf{x}_t^m \in \mathbb{R}^D\) contains features such as temperature, GDP, and population indicators at month \(t\).

\textbf{Input Embedding.}
We first project each \(\mathbf{x}_t^m\) into a \(d_{\text{model}}\)-dimensional space and add a learnable positional encoding \(\mathbf{p}_t \in \mathbb{R}^{d_{\text{model}}}\) to encode the temporal order of months. Specifically,
\begin{equation}
  \mathbf{e}_t^m 
  \;=\;
  \mathrm{Linear}\!\bigl(\mathbf{x}_t^m\bigr) 
  \;+\;
  \mathbf{p}_t,
  \label{eq:embedding}
\end{equation}
where \(\mathrm{Linear}(\cdot)\) is a fully connected layer mapping \(\mathbb{R}^D\rightarrow \mathbb{R}^{d_{\text{model}}}\). By concatenating embeddings across all \(t\in\{1,\dots,T\}\), we obtain
\begin{equation}
  \mathbf{E}^m 
  \;=\; 
  \bigl[\mathbf{e}_1^m;\,\mathbf{e}_2^m;\,\dots;\,\mathbf{e}_T^m\bigr]
  \;\in\; 
  \mathbb{R}^{T \times d_{\text{model}}}.
  \label{eq:embedding_matrix}
\end{equation}

\textbf{Masked Multi-Head Attention Encoder.}
We feed \(\mathbf{E}^m\) into three stacked Transformer encoder layers, each consisting of:
\begin{equation}
  \mathbf{Z}^{(l)} 
  \;\xrightarrow{\mathrm{MHA}}\; 
  \mathbf{Z}'^{(l)}
  \;\xrightarrow{\mathrm{FFN}}\; 
  \mathbf{Z}^{(l+1)}, 
\end{equation}
where \(l\in\{1,\dots,h]L\}\). Let \(\mathbf{Z}^{(1)} = \mathbf{E}^m\). Each layer comprises:

\begin{itemize}
  \item Masked Multi-Head Attention (MHA). 
  Define
  \begin{equation}
    \mathbf{Q} = \mathbf{Z}^{(l)} \mathbf{W}_{Q}, 
    \quad
    \mathbf{K} = \mathbf{Z}^{(l)} \mathbf{W}_{K},
    \quad
    \mathbf{V} = \mathbf{Z}^{(l)} \mathbf{W}_{V},
    \label{eq:QKV}
  \end{equation}
  where
  \(\mathbf{W}_{Q}, \mathbf{W}_{K}, \mathbf{W}_{V} \in \mathbb{R}^{d_{\text{model}} \times d_k}\),
  \(d_k = d_{\text{model}} / h\), and \(h=4\) is the number of attention heads. For each head \(i\in\{1,\dots,h\}\),
  \begin{equation}
    \mathrm{head}_i \;=\;
    \mathrm{Softmax}\!\Bigl(
      \mathrm{Mask}\!\bigl(
        \tfrac{\mathbf{Q}_i \mathbf{K}_i^\top}{\sqrt{d_k}}
      \bigr)
    \Bigr)\,
    \mathbf{V}_i,
    \label{eq:head}
  \end{equation}
  where \(\mathrm{Mask}(\cdot)\) ensures causal attention (i.e., time step \(t\) cannot attend to \(t'>t\)). Concatenating the heads, we get
  \begin{equation}
    \mathrm{MHA}(\mathbf{Z}^{(l)}) \;=\;
    \mathrm{Concat}\bigl(\mathrm{head}_1,\dots,\mathrm{head}_h\bigr)
    \,\mathbf{W}_O,
    \label{eq:mha_concat}
  \end{equation}
  with \(\mathbf{W}_O \in \mathbb{R}^{(h\,d_k)\times d_{\text{model}}}\).

  \item Position-Wise Feed-Forward Network (FFN).
  Each encoder layer also includes a two-layer Multi-layer Perceptron(MLP) with a ReLU activation:
  \begin{equation}
    \mathrm{FFN}(\mathbf{z}) 
    \;=\;
    \max\!\bigl(\mathbf{0}, \mathbf{z}\,\mathbf{W}_1 + \mathbf{b}_1\bigr)
    \,\mathbf{W}_2 + \mathbf{b}_2,
    \label{eq:ffn}
  \end{equation}
  wrapped with residual connections and layer normalization:
  \begin{equation}
    \mathbf{Z}'^{(l)} 
    \;=\; 
    \mathrm{LayerNorm}\!\Bigl(
      \mathbf{Z}^{(l)} + \mathrm{MHA}\!\bigl(\mathbf{Z}^{(l)}\bigr)
    \Bigr),
  \end{equation}
  \begin{equation}
    \mathbf{Z}^{(l+1)} 
    \;=\; 
    \mathrm{LayerNorm}\!\Bigl(
      \mathbf{Z}'^{(l)} 
      + \mathrm{FFN}\!\bigl(\mathbf{Z}'^{(l)}\bigr)
    \Bigr).
  \end{equation}
\end{itemize}
In our implementation, \(d_{\text{model}}=256\) , the FFN inner dimension \(d_{\text{ff}}=512\) and layer number of the attention encoder $L$ is set to 3.

\textbf{Final Time-Step Representation and Classification.}
After three encoder layers, the output
\(\mathbf{Z}^{(L)} \in \mathbb{R}^{T\times d_{\text{model}}}\)
contains context-rich representations of each month. We extract
\(\mathbf{z}_T^m \in \mathbb{R}^{d_{\text{model}}}\),
the representation at the final time step \(T\), and feed it into a three-layer MLP (hidden dimension \(256\)), yielding a 2-dimensional logit vector \(\mathbf{o}^m \in \mathbb{R}^2\):
\begin{equation}
  \mathbf{o}^m 
  \;=\; 
  \mathrm{MLP}\!\bigl(\mathbf{z}_T^m\bigr),
  \label{eq:logits}
\end{equation}
which supports a binary prediction regarding whether city \(m\) is at high or low risk in the next month \((T+1)\). Further details concerning loss functions and optimization are provided in Section~\ref{sec:training}.

\subsection{Training Process}\label{sec:training}

Considering the limited number of definitively labeled samples and the presence of \emph{uncertain} data (whose underlying true risks are inferred only probabilistically via Wilson intervals), we adopt a multi-task training paradigm. Specifically, we first perform three \emph{pretraining} tasks---Time-Series Self-Supervised Learning, City Safety Ranking, and City Risk Ranking---to maximize the model’s exposure to supervisory signals. We then carry out a \emph{semi-supervised} training step, leveraging Wilson-based soft labels to refine the final classification of city-level food safety risks.

\textbf{Time-Series Self-Supervised Learning.}
In this stage, the goal is to predict each city’s future features from its past observations, thereby learning robust temporal representations. For each city \(m\) with feature sequence
\(\{\mathbf{x}_1^m,\dots,\mathbf{x}_T^m\}\), the Transformer encoder (detailed in Section~\ref{sec:model}) produces a sequence of hidden states 
\(\{\mathbf{h}_1^m,\dots,\mathbf{h}_{T-1}^m\}\). We introduce a dedicated MLP, denoted \(g(\cdot)\), to predict \(\mathbf{x}_{t+1}^m\) from \(\mathbf{h}_t^m\):
\begin{equation}
  \widehat{\mathbf{x}}_{t+1}^m 
  \;=\;
  g\!\Bigl(\mathbf{h}_t^m\Bigr),
  \quad
  t \;=\; 1,\dots,T-1.
  \label{eq:ss_pred}
\end{equation}
The training objective is to minimize the discrepancy between the predicted features and the ground-truth next-step features, typically measured by a mean-squared error (MSE):
\begin{equation}
  \mathcal{L}_\text{self}
  \;=\;
  \sum_{m=1}^{M}
  \sum_{t=1}^{T-1}
  \bigl\|
    \widehat{\mathbf{x}}_{t+1}^m 
    \;-\;
    \mathbf{x}_{t+1}^m
  \bigr\|^2.
  \label{eq:ss_loss}
\end{equation}
By constructing multiple \((\text{input},\text{target})\) pairs from a single time series, this task substantially increases the effective training set and fosters the encoder’s ability to extract informative spatiotemporal patterns.

\textbf{City Safety Ranking Learning.}
Following self-supervised pretraining, we incorporate partially supervised signals derived from \emph{passing} (\ie~compliant) test results. Based on the Wilson interval, each city’s \emph{safety} ranking can be represented by a continuous score \(s_m^\text{safe}\), normalized to \([0,1]\) (where 1 signifies the highest failure rate, 0 the lowest). We first obtain each city’s final hidden representation \(\mathbf{z}_T^m\) (Eq.~\eqref{eq:logits}), then compute an aggregated embedding
\begin{equation}
  \mathbf{u}_m^\text{safe}
  \;=\;
  \bigl[
    \mathbf{z}_T^m
    \;\Vert\;
    \mathrm{Avg}\!\bigl(
      \{
        \mathbf{z}_T^n \mid n \neq m
      \}
    \bigr)
  \bigr],
  \label{eq:ranking_embedding_safe}
\end{equation}
where \(\Vert\) denotes concatenation and \(\mathrm{Avg}(\cdot)\) computes the mean of other cities’ embeddings. A safety-ranking MLP, denoted \(r_\text{safe}(\cdot)\), then maps \(\mathbf{u}_m^\text{safe}\) into a scalar \(\widehat{s}_m^\text{safe}\in [0,1]\). We train via MSE:
\begin{equation}
  \mathcal{L}_\text{safe}
  \;=\;
  \sum_{m=1}^{M}
  \Bigl(
    \widehat{s}_m^\text{safe}
    \;-\;
    s_m^\text{safe}
  \Bigr)^{2},
  \label{eq:ranking_loss_safe}
\end{equation}
enforcing that the model’s predicted safety scores align with the Wilson-based ranking.

\textbf{City Risk Ranking Learning.}
An analogous ranking task is designed for \emph{failing} test results. Let \(s_m^\text{risk}\in [0,1]\) be the Wilson score-based \emph{risk} ranking for city \(m\), where a value near 1 indicates a higher estimated failure rate. Similar to Eq.~\eqref{eq:ranking_embedding_safe}, we form 
\(\mathbf{u}_m^\text{risk}\) by concatenating \(\mathbf{z}_T^m\) with the average of other cities’ embeddings. A distinct MLP, \(r_\text{risk}(\cdot)\), produces \(\widehat{s}_m^\text{(risk)}\in [0,1]\):
\begin{equation}
  \mathcal{L}_\text{risk}
  \;=\;
  \sum_{m=1}^{M}
  \Bigl(
    \widehat{s}_m^\text{risk}
    \;-\;
    s_m^\text{risk}
  \Bigr)^{2}.
  \label{eq:ranking_loss_risk}
\end{equation}
Training on both \(\mathcal{L}_\text{safe}\) and \(\mathcal{L}_\text{risk}\) helps the model differentiate cities with truly high compliance rates from those prone to food safety incidents.

\textbf{Semi-Supervised City Risk Prediction.}
Following the three pretraining tasks, we fine-tune the \emph{total classification classfication model} (described in Section~\ref{sec:model}) to judge whether each city \(m\) is at high or low risk in the subsequent month \((T+1)\). While certain city-month samples have \emph{definitive} labels (e.g., when their Wilson interval lower bound lies above a predefined threshold \(\tau\)), many are \emph{uncertain} due to wide confidence intervals. To leverage these uncertain samples, we assign soft labels via:
\begin{equation}
  p_m
  \;=\;
  \mathrm{Prob}\!\bigl(
    \text{failure rate} \;\ge\; \tau
    \,\big|\,
    L_m,\,U_m
  \bigr),
  \label{eq:p_m}
\end{equation}
where \([L_m, U_m]\) is the Wilson interval for city \(m\)’s failure rate. Interpreting \(p_m\) as the probability of high risk, each city’s two-class label becomes 
\(\bigl(1 - p_m,\, p_m\bigr)\). Let 
\(\mathbf{o}^m \in \mathbb{R}^2\) be the logits produced by the classification MLP when given \(\mathbf{z}_T^m\), the final hidden representation from the Transformer encoder:
\begin{equation}
  \mathcal{L}_\text{semi}
  \;=\;
  -\sum_{m=1}^{M}\,
  \Bigl[
    (1 - p_m)\,\log\!\Bigl(\sigma\!\bigl(\mathbf{o}^m\bigr)_\text{low}\Bigr)
    \;+\;
    p_m\,\log\!\Bigl(\sigma\!\bigl(\mathbf{o}^m\bigr)_\text{high}\Bigr)
  \Bigr],
  \label{eq:semi_loss}
\end{equation}
where \(\sigma(\cdot)\) denotes the softmax function. This semi-supervised approach augments the limited pool of definitively labeled examples with probabilistic labels for uncertain samples, improving the overall robustness of city-level risk prediction.

In summary, four separate MLP heads (\ie~\(g(\cdot)\), \(r_\text{safe}(\cdot)\), \(r_\text{risk}(\cdot)\), and the final risk classifier) share the same Transformer encoder. By first performing self-supervised feature prediction and city-level ranking tasks, we derive meaningful representations of each city’s food safety profile. The subsequent semi-supervised step then refines these representations, yielding robust performance in city-level risk prediction even when a large fraction of the data are uncertain.

\subsection{Experiment Setting}
\label{sec:experiment_setting}

\textbf{Training and Testing.}
To rigorously evaluate our framework, we partition the data at the end of 2021. All city-month observations collected before 2022 are used for a 10-fold cross validation scheme, generating ten independently trained models and corresponding validation sets. Specifically, each fold alternates which 10\% of pre-2022 city-month samples serve as validation data, while the remaining 90\% act as training data. This procedure provides a more robust estimation of generalization performance by reducing overfitting to any single partition of the dataset.

We then use the ten trained models to predict on city-month samples from 2022 onward. Since actual food safety outcomes during these later months can be partially uncertain, we exclude city-months with \textit{ambiguous} Wilson intervals (i.e., those that do not confidently indicate either high or low risk). Restricting our test set to definitively labeled cases ensures a clearer measure of predictive accuracy. Averaging the results across the ten model predictions yields a final performance assessment that captures both within-sample variability (from cross validation) and out-of-sample validation (from 2022). 

\textbf{Implementation Details.}  
We implement the deep learning models in Python using PyTorch.
Training is conducted on an NVIDIA A800 GPU, running for a maximum of 100 epochs unless early stopping criteria are met. We use the Adam optimizer (learning rate set to \(10^{-4}\), \(\beta_1=0.9\), \(\beta_2=0.999\)) with a mini-batch size of 16. 
For each of the three pretraining tasks (Section~\ref{sec:training}), we warm up the model over 20 epochs to stabilize convergence, subsequently fine-tuning on the semi-supervised city risk classification for another 30--50 epochs, based on validation loss. 
Early stopping is applied if the validation performance does not improve for 10 consecutive epochs; in such a case, the model parameters yielding the best validation loss are retained.

Upon completion of training, we report the model’s performance on the test set, focusing on accuracy, F\(_1\)-score, precision, recall, and specificity metrics detailed in our results(Section~\ref{sec:Numerical experiment}).

\bibliography{sections/sn-bibliography}

\end{document}